\documentclass{article}
\pdfoutput=1

     \usepackage[final]{neurips_2019}

\usepackage[utf8]{inputenc} 
\usepackage[T1]{fontenc}    
\usepackage{hyperref}       
\usepackage{url}            
\usepackage{booktabs}       
\usepackage{amsfonts}       
\usepackage{nicefrac}       
\usepackage{microtype}      

\usepackage{graphicx}
\usepackage{subcaption}

\title{Helping Reduce Environmental Impact of Aviation with Machine Learning}

\author{%
  Ashish Kapoor \\
  Microsoft Corporation\\
  1 Microsoft Way\\
  Redmond, WA 98052 \\
  \texttt{akapoor@microsoft.com} \\
}

\begin{document}

\maketitle

\begin{abstract}
Commercial aviation is one of the biggest contributors towards climate change. We propose to reduce environmental impact of aviation by considering solutions that would reduce the flight time. Specifically, we first consider improving winds aloft forecast so that flight planners could use better information to find routes that are efficient. Secondly, we propose an aircraft routing method that seeks to find the fastest route to the destination by considering uncertainty in the wind forecasts and then optimally trading-off between exploration and exploitation. Both these ideas were previously published in \cite{kapoor2014airplanes} and \cite{Sun2016NoregretRU} and contain further technical details.
\end{abstract}

\section{Introduction}
Commercial aviation has a severe effect on our climate due to emission of harmful particulates and gases such as carbon dioxide, carbon monoxide, hydrocarbons, nitrogen oxides, sulfur oxides, lead etc. In 2017, aviation resulted in 859 million tonnes of CO2, which is roughly 2\% of man-made carbon emissions \cite{acp-10-6391-2010}. According to some estimates, one round-trip of trans-Atlantic flight emits enough carbon dioxide to melt 30 square feet of Arctic sea ice. Consequently, it is important to seriously consider efforts that promise to limit the adverse effects of aviation. While there are joint efforts in the industry such as Carbon Offsetting and Reduction Scheme for International Aviation (CORSIA)\cite{corsia}, they are mostly voluntary, market based and have modest targets. 

We explore and propose ML methods to optimize for flight time given a source and a destination. The key insight is an aircraft can significantly reduce flight time by choosing routing that has favorable winds. Consequently, our proposal is two fold: use machine learning to (1) improve winds aloft forecast and (2) derive flight policies that are optimal for time to reach destination. We propose to use the network of aircraft already in the air as surrogate sensors that continually update and inform about the winds. We also explore the idea of sequential decision making via contextual bandits to derive optimal flight policies. Shaving an hour of flight time on average can reduce carbon dioxide emissions by over 2500 kg, hence the methods described in this paper can be fairly useful.

\begin{figure}[t]
    \centering
    \includegraphics[width=\textwidth]{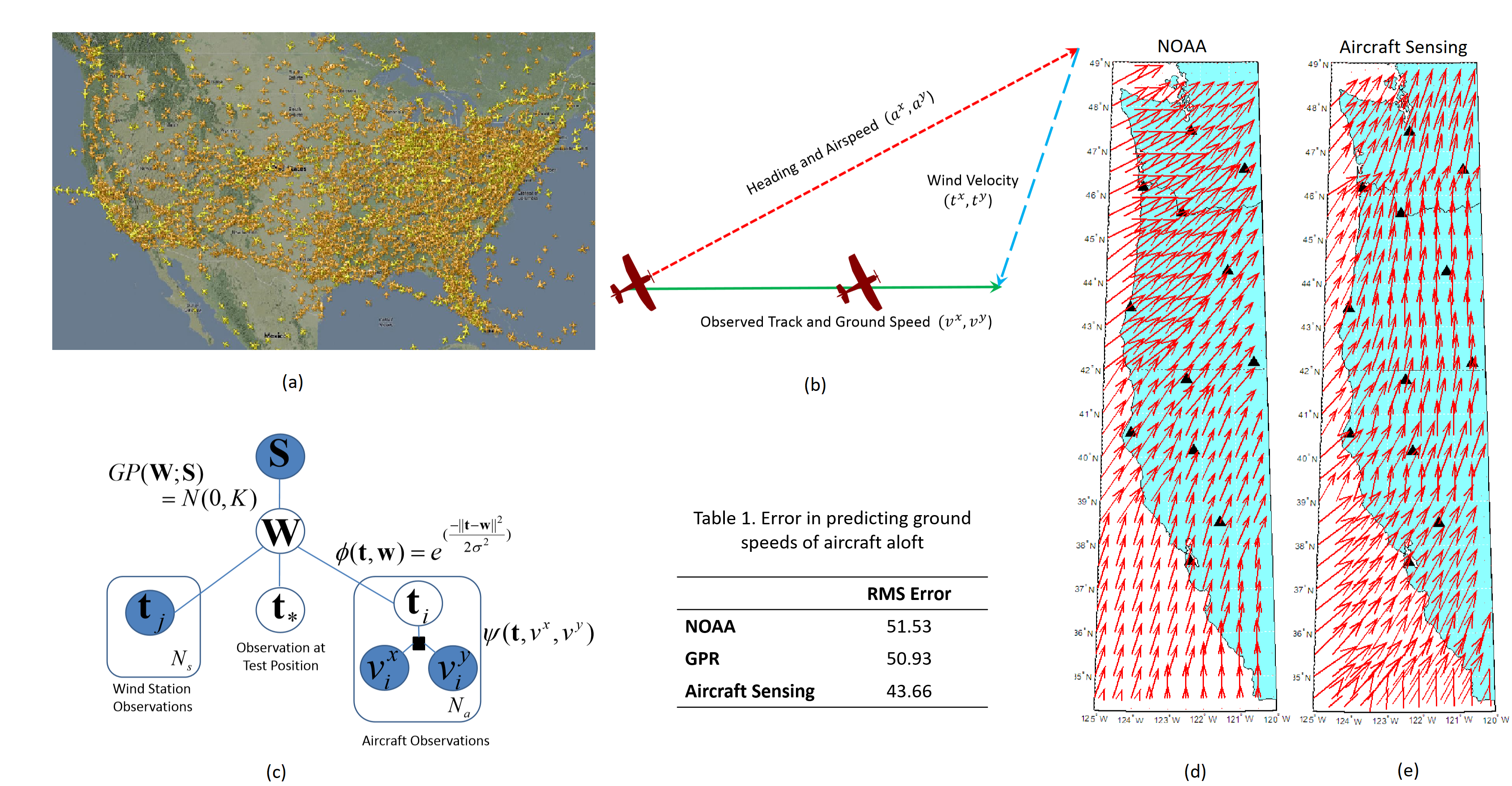}
    \caption{(a) Visualisation of data on all flying aircraft over the continental US.
    (b) Graphical representation of the wind triangle. The aircraft maintains an assigned heading at a reported constant and known airspeed $(a^{x},a^{y})$, while the observed wind $(t^{x},t^{y})$ changes the observed course and the ground speed $(v^x,v^y)$ via a vector addition.
    (c) The probabilistic graphical model for wind prediction. (d) and (e)  Comparison of estimated winds at 30K feet with the different models.
    Table 1. shows the RMS error in predicting aircraft ground speeds indicating the advantage of our method.}
    \label{fig:concept}
    \vspace{-0.2in}
\end{figure}

\subsection{Improving Wind Forecasts}
The core idea is to use publicly available information from the large number of aircraft aloft (fig. \ref{fig:concept}(a)) to improve wind forecasts. US National Oceanic and Atmospheric Administration (NOAA) publishes Winds Aloft report for static wind station sites at a discrete set of altitudes. We refine these forecasts by considering how fast an aircraft is traveling over the ground (ground speed), which is highly informative about the latent winds that an aircraft is experiencing at that location (fig. \ref{fig:concept}(b)).

Formally, we use Gaussian Processes (GPs)\cite{rasmussen2006gaussian} as they provide an efficient updating procedure for conditioning phenomena at one location based on observations at other locations. Fig. \ref{fig:concept}(c) shows the graphical model corresponding to the proposed approach, where ${\bf W}$ denotes the true winds which are latent. The model shows the observed quantities as shaded nodes, i.e., the collection of sites ${\bf S}$, observed winds at wind stations ${\bf T}_L$ and the observed over the ground velocities ${\bf V}_A = [{\bf v}_i]_{i = 1}^{N_A}$ for the aircraft. The boxes represent repetitions: $N_s$ for the different wind stations and $N_a$ for the separate aircraft.
GPs allow us to encode that similar winds are observed at nearby sites, and incorporate additional information from the aircraft aloft.
Formally, it combines the smoothness constraint and the information obtained by observations ${\bf T}_L$ and ${\bf V}_A$ at the wind stations and the aircraft
respectively. For the wind stations, the probabilistic relationship between the hidden true wind ${\bf w}$ and the noisy observation ${\bf t}$
is defined via a Gaussian likelihood model $\phi({\bf t}, {\bf w}): = e^{-\frac{||{\bf t} - {\bf w}||^2}{2\sigma^2}}$.

Accessing data about winds from the airplanes in flight requires an extra step as the public reports from the planes provide only the ground velocities of the aircraft. We identify winds from ground velocity reports via the potential function $\psi({\bf t}, v^x, v^y) = e^{-\beta(||{\bf v}-{\bf t}||-||{\bf a}||)^2}$ that relates the ground velocity of aircraft to the encountered winds via the wind triangle (Figure \ref{fig:concept}(b)). In summary, the proposed model ties the site locations ${\bf S}$, the
true wind ${\bf W}$, the noisy wind observations ${\bf t}$, and the aircraft ground velocities ${\bf v}$ by inducing the following distribution:
$p({\bf T}, {\bf V}_{A}, {\bf W} | {\bf S}) \propto GP({\bf W}; {\bf S})\times\prod_{i \in L \cup A \cup *}{\phi({\bf t}_i, {\bf w}_i)} \prod_{j \in A}{\psi({\bf t}_j, v^x_j, v^y_j)}.$
Here ${\bf T} = {\bf T}_L \cup {\bf T}_A \cup {\bf t}_*$. We use Laplace approximation \cite{Laplace} to infer the winds ${\bf W}$, which results in a refinement of the NOAA forecasts.

\subsection{Exploiting Wind Forecast for Efficient Routing}
Next, we propose a routing policy for aircraft that is sensitive to the prevailing winds and optimizes for the travel time. We consider an online receding horizon based path planner \cite{urmson2008autonomous} that operates in an environment with latent information modelled as a Gaussian Process. Equipped with a pre-computed trajectory library $\{\tau_1,...,\tau_K\}$, at every iteration our algorithm picks a trajectory to execute while collecting observations of the latent wind on the fly to update the GP. Corresponding to each trajectory $\tau_i$ there is a reward function $f_{i}$, which depends on the uncertain winds ${\bf W}$. The reward reflects how quickly is the aircraft moving towards the destination. The ideal goal of the aircraft is to pick a sequence of trajectories, so that it can maximize the cumulative reward. It is not easy for the aircraft to pre-plan a sequence of the decisions since the aircraft does not have the exact information about the winds except a prior through winds aloft reports. Hence the aircraft needs to plan on the fly while collecting new information and refining its knowledge about the winds for future planning. 

Our method UCB-Replanning leverages the idea of optimism in the face of uncertainty to tradeoff exploration and exploitation in a near-optimal manner. We design an algorithm similar to GP-UCB (Upper Confidence Bound) \cite{srinivas2009gaussian}, where we maintain a confidence interval of the true reward for each trajectory at every step. This is accomplished by first extracting confidence intervals of the uncertain winds ${\bf W}$ from GP and then using them to compute the variance in reward for each trajectory. We traverse a trajectory with the highest upper confidence bound of the reward. The new measurements are then used to refine the wind predictions ${\bf W}$ and the procedure is repeated till the destination.

\section{Experiments and Results}
\noindent{\bf Improving Wind Forecasts: }We first logged data from aircraft flying over the states of WA, OR, and CA using flightaware. The data consisted of $1653$ observations from $496$ unique aircraft. We explore two additional baselines 1) nearest-neighbor wind interpolation using NOAA forecasts, which is similar to existing flight planners \cite{jeppesen2001pri} and 2) non-linear interpolation via GP regression (GPR). Fig. \ref{fig:concept} (d) and (e) graphically shows wind predictions at 30K feet and we observe that the estimated winds by the proposed method are significantly different than the NOAA forecasts. Beyond the qualitative difference in forecasts, we empirically evaluate correctness of the methods by predicting the true ground speed of flying aircraft given wind forecasts. We use leave-one-aircraft-out methodology, where the predictive model is built on all but one test aircraft, and the procedure repeated by considering each airplanes as a test data. Table 1. shows that the forecast that uses aircraft data provides better root mean squared error ($43.66$) than others ($51.53$ for NOAA, $50.93$ for GPR). 

\begin{figure}[t]
   \begin{minipage}{0.66\textwidth}
	\begin{subfigure}[l]{0.49\textwidth}
        \includegraphics[trim=20mm 20mm 50mm 20mm, clip, width=1.0\textwidth]{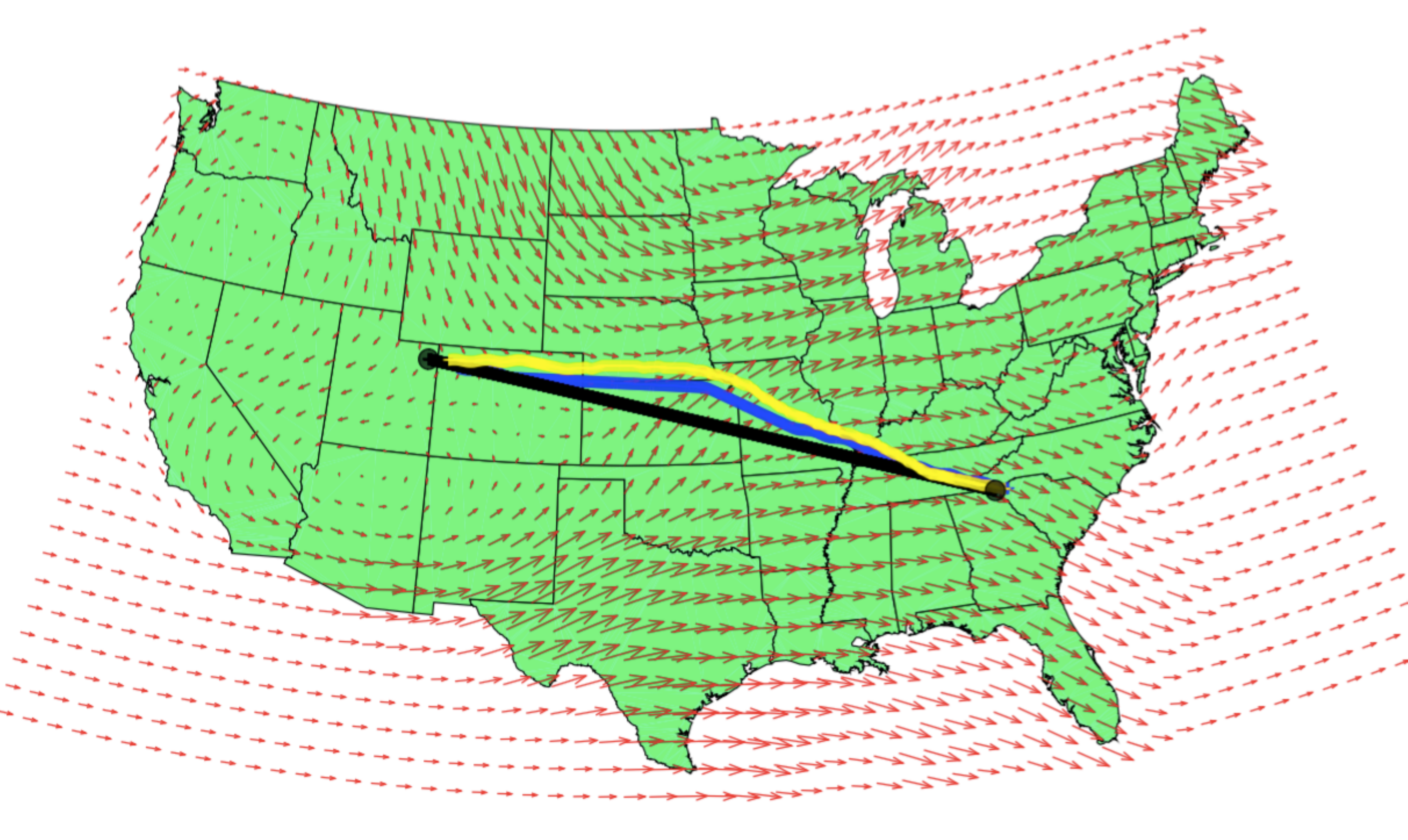}
    	\caption{SC to UT (Head wind)}
    \end{subfigure}
  	\begin{subfigure}[l]{0.49\textwidth}
        \includegraphics[trim=20mm 20mm 50mm 20mm, clip, width=1.0\textwidth]{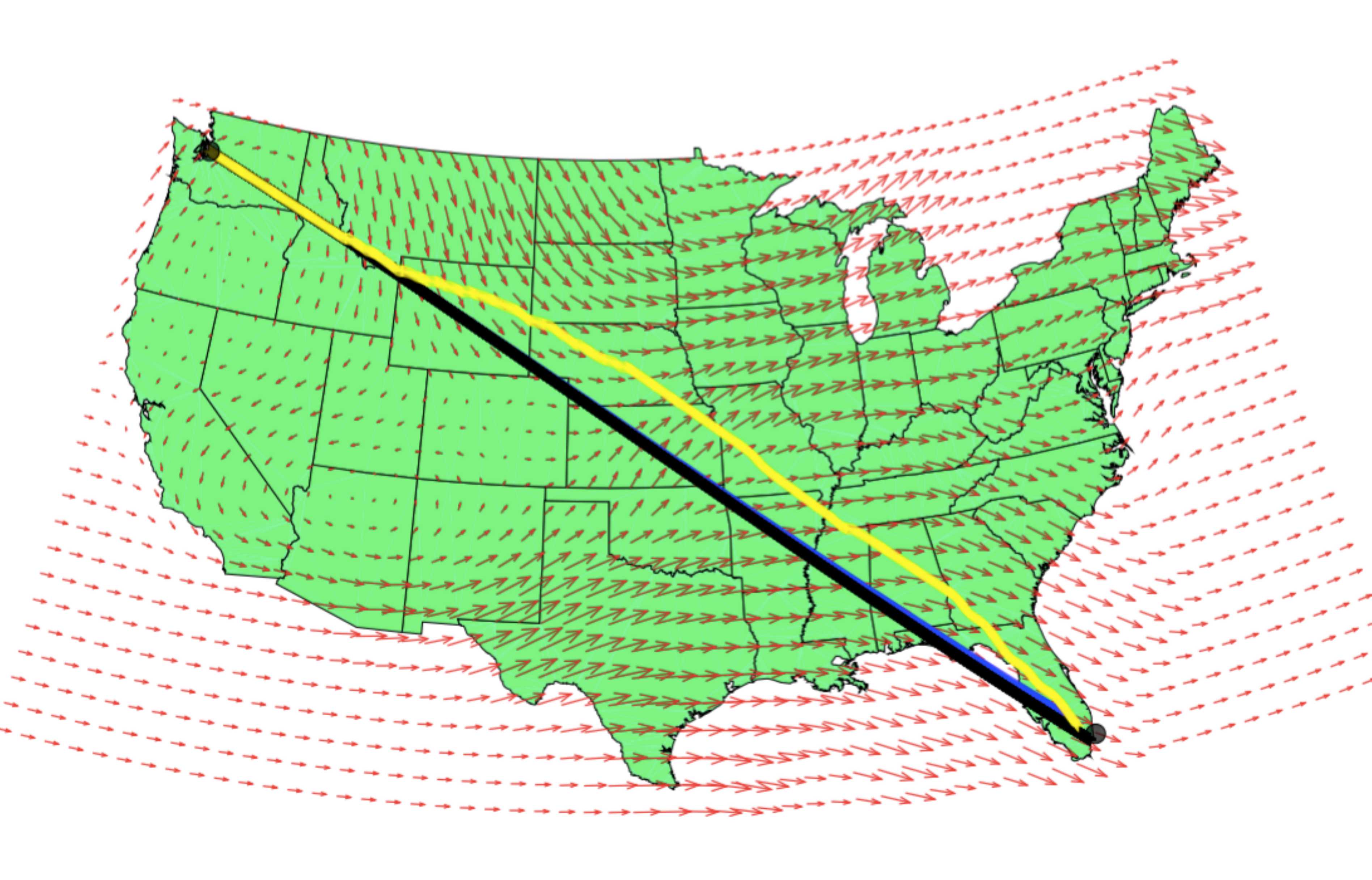}
    	\caption{Seattle to Miami (Tail Wind)}
    \end{subfigure}
   \end{minipage}
   \begin{minipage}{0.33\textwidth}
\begin{center}
   {\scriptsize
   Table 2: Average Travel Times in Seconds\\
   \begin{tabular}{llr}
   \toprule
    & SC to UT   & Seattle to Miami \\ 
    \midrule
    \textbf{UCB} & \textbf{21079.7}$\pm$1109 & \textbf{31333.1}$\pm$1269  \\ 
    \textbf{Mean} &  21183.3$\pm$1263 & {31716.5$\pm$1016} \\ 
    \textbf{GCR} & 33712.5$\pm$1852 &  48195.7$\pm$1952 \\ 
    \bottomrule
    \end{tabular}}
\end{center}
\label{tab:comparison_real}
\end{minipage}
\caption{Trajectories from UCB (blue), Mean (yellow) and Great Circle (black) for a route from SC to UT with a head wind (a), and a route from Seattle to Miami with a tail wind (b). Table 2 shows average traveling time (in seconds) with standard deviation for different methods with real winds.}
\label{fig:real_wind_map}
\vspace{-0.2in}
\end{figure}

\noindent{\bf Improving Routing: }Our experiments simulate an aircraft that maintains a constant cruising speed of 250 knots at an altitude of 39000 feet (11887 m). The winds aloft forecast provided by NOAA is used to construct a ground truth wind map via a GP. 
We use an existing pre-computed library of trajectories from \cite{green2007toward}. We test UCB (proposed method) and Mean (when just using mean prediction to choose best trajectory) on two different routes: (1) a \emph{short} route from South Carolina to Utah (around 1300 nautical miles),  and (2) a \emph{long} route from Seattle to Miami (around 2700 nautical miles). Fig.~\ref{fig:real_wind_map} (a) shows that when flying with head wind, that UCB and Mean saves traveling time by flying in the direction that is nearly perpendicular to the wind speed in order to cancel the wind effect when the wind is strong. However, when flying with good tail wind as in Fig.~\ref{fig:real_wind_map} (b)), UCB almost follows with Great Circle Route. We repeat this experiment over 11 days by dividing each day into 6 hour time slots and simulating both paths for every slot. We report the average traveling time and std. deviation in Table 2. Results show that UCB has significant advantages over others. 

\section{Conclusion and Future Work}
Commercial aviation is one of the biggest contributors to climate change. We first propose to improve wind aloft forecast, using data from the network of aircraft already in the air, so that most up-to-date information could be used for more efficient flight plans. Secondly we look into flight routing policies that can dynamically decide on the most efficient route to take. The proposal to improve wind forecasts have better chances of implementation as the existing flight planners can simply point to the refined forecasts. The second approach will require significant changes in regulations and policies before it gets implemented. We welcome the readers to access our online service with refined wind forecasts for continental US at {\it anonymous}. 
\bibliographystyle{abbrv}
\bibliography{allbib}

\begin{thebibliography}{10}

\bibitem{Laplace}
A.~Azevedo-Filho and R.~Shachter.
\newblock Laplace's method approximations for probabilistic inference in belief
  networks with continuous variables.
\newblock In {\em Uncertainty in Artificial Intelligence}, 1994.

\bibitem{green2007toward}
C.~Green and A.~Kelly.
\newblock Toward optimal sampling in the space of paths.
\newblock In {\em 13th International Symposium of Robotics Research}, 2007.

\bibitem{corsia}
IATA.
\newblock Fact sheet: Climate change and {C}{O}{R}{S}{I}{A}, 2017.

\bibitem{jeppesen2001pri}
Jeppsen.
\newblock {\em Private Pilot Manual}.
\newblock Sanderson, 2001.

\bibitem{kapoor2014airplanes}
A.~Kapoor, Z.~Horvitz, S.~Laube, and E.~Horvitz.
\newblock Airplanes aloft as a sensor network for wind forecasting.
\newblock In {\em Proceedings of the 13th international symposium on
  Information processing in sensor networks}, pages 25--34. IEEE Press, 2014.

\bibitem{rasmussen2006gaussian}
C.~E. Rasmussen.
\newblock {\em Gaussian processes for machine learning}.
\newblock Citeseer, 2006.

\bibitem{srinivas2009gaussian}
N.~Srinivas, A.~Krause, S.~M. Kakade, and M.~Seeger.
\newblock Gaussian process optimization in the bandit setting: No regret and
  experimental design.
\newblock {\em arXiv preprint arXiv:0912.3995}, 2009.

\bibitem{Sun2016NoregretRU}
W.~Sun, N.~Sood, D.~Dey, G.~Ranade, S.~Prakash, and A.~Kapoor.
\newblock No-regret replanning under uncertainty.
\newblock {\em 2017 IEEE International Conference on Robotics and Automation
  (ICRA)}, 2016.

\bibitem{urmson2008autonomous}
C.~Urmson, J.~Anhalt, D.~Bagnell, C.~Baker, R.~Bittner, M.~Clark, J.~Dolan,
  D.~Duggins, T.~Galatali, C.~Geyer, et~al.
\newblock Autonomous driving in urban environments: Boss and the urban
  challenge.
\newblock {\em Journal of Field Robotics}, 25(8):425--466, 2008.

\bibitem{acp-10-6391-2010}
J.~T. Wilkerson, M.~Z. Jacobson, A.~Malwitz, S.~Balasubramanian, R.~Wayson,
  G.~Fleming, A.~D. Naiman, and S.~K. Lele.
\newblock Analysis of emission data from global commercial aviation: 2004 and
  2006.
\newblock {\em Atmospheric Chemistry and Physics}, 10(13), 2010.

\end{thebibliography}

\end{document}